\documentclass{bmvc2k}

\title{MoBYv2AL: Self-supervised Active Learning for Image Classification}

\addauthor{Razvan Caramalau}{r.caramalau18@imperial.ac.uk}{1}
\addauthor{Binod Bhattarai}{b.bhattarai@ucl.ac.uk}{2}
\addauthor{Danail Stoyanov}{danail.stoyanov@ucl.ac.uk}{2}
\addauthor{Tae-Kyun Kim}{tk.kim@imperial.ac.uk}{1,3}

\addinstitution{
 Imperial College London, UK
}
\addinstitution{
 University College London, UK
}
\addinstitution{
 School of Computing,\\
 KAIST, 
 Daejeon, South Korea
}
\runninghead{CARAMALAU ET AL.}{MoBYv2AL: Self-supervised Active Learning}

\usepackage{graphicx}
\begin{document}

\maketitle

\begin{abstract}
Active learning(AL) has recently gained popularity for deep learning(DL) models. This is due to efficient and informative sampling, especially when the learner requires large-scale labelled datasets. Commonly, the sampling and training happen in stages while more batches are added. One main bottleneck in this strategy is the narrow representation learned by the model that affects the overall AL selection.

We present \emph{MoBYv2AL}, a novel self-supervised active learning framework for image classification. Our contribution lies in lifting MoBY -- one of the most successful self-supervised learning algorithms to the AL pipeline. Thus, we add the downstream task-aware objective function and optimize it jointly with contrastive loss. Further, we derive a data-distribution selection function from labelling the new examples. Finally, we test and study our pipeline robustness and performance for image classification tasks. We successfully achieved state-of-the-art results when compared to recent AL methods. Code available: \url{https://github.com/razvancaramalau/MoBYv2AL}
\end{abstract}

\section{Introduction}
\label{sec:intro}
Active Learning (AL)~\cite{Sinha2019VariationalLearning,ugcn,Yoo2019LearningLearning,KimConfidentAnalysis, cdal,csal, bengar2021reducing,kim2021task} has recently gained more popularity in the research community. The goal of AL is to sample the most \emph{informative} and \emph{diverse} examples from a large pool of unlabelled data to query their labels. The existing AL methods can be grouped into two based on the selection criteria. The first group is
uncertainty-based algorithms~\cite{mcdropoutal,Yoo2019LearningLearning,Gal2016DropoutGhahramani} that select the challenging and informative examples. Whereas representative-based algorithms select the most diverse examples from the data set. To select diverse examples, existing methods first project the images into a feature space followed by applying sampling techniques such as CoreSet~\cite{Sener2017ActiveApproach}. 
Our work falls in the latter category.  

Prominent works on representative-based methods for AL in the past few years have tackled 
a wide range of architectures to learn the image representations such as Convolutional Neural Network~\cite{Yoo2019LearningLearning}, Graph Convolutional  Neural Networks~\cite{ugcn}, Bayesian Network~\cite{caramalau2021active}, Variational Auto-Encoders~\cite{Sinha2019VariationalLearning,kim2021task}, 
and too few to mention. These works have proven that the learned features of the images have directly 
influenced the performance of the pipeline. However, these methods suffer from \emph{cold-start problem}. As we know, in the early selection stage, we have limited annotated examples, and the above-mentioned architectures
are hard to train with the small number of training examples. Thus, the features extracted from such models get biased from the beginning and continue to become sub-optimal in the subsequent selection stages. This problem is commonly known as \emph{cold-start problem}. To address such a problem, recent works in AL have explored self-supervised learning methods~\cite{csal, bengar2021reducing, altod, revival_iccv21}.

Self-supervised learning methods~\cite{mixmatch,xie2021moby,dino,moco,byol} have made tremendous progress in generating discriminative representations of the images. Some methods have even come close to supervised methods in generalization~\cite{chen2020mocov2,byol,xie2021moby}. One of the earliest works in this direction~\cite{csal} employed consistency loss between the input image and its geometrically augmented versions along with the objective of downstream tasks. However, this method limits augmentation methods in the primitive form. Similarly, J. Bengar et al. ~\cite{bengar2021reducing} introduced contrastive learning in AL, but the self-supervised method and end-task objective are optimised in multi-stage form. This makes the model sub-optimal, affecting the features' representativeness during selection. Simple random labelling overpasses any AL criteria. Thus, the existing works in this direction show explicit limitations. 

To address the issues of those methods, we introduce contrastive learning as MoBYv2 (from its predecessor MoBY~\cite{xie2021moby})  in our AL pipeline, \emph{MoBYv2AL}, and jointly train the learner. We choose  MoBY SSL because it addresses the computational complexities and shortcomings of other previous methods, such as SimCLR~\cite{simclr} or BYOL\cite{byol}. MoBY
has two branches (as shown in Figure~\ref{fig:pipeline}). One updates with gradient (query encoder) and 
another with momentum (key encoder). The parameters of the momentum encoder are updated in slow-moving averages with the query one. Moreover, the memory bank of keys from the momentum encoder keeps long dependencies with several mini-batches. Apart from minimising a contrastive loss, another advantage consists in the asymmetric structure of BYOL that captures distances from mean representation. The AL process of MoBYv2AL culminates with the concept-aware selection function, CoreSet.

We state our contributions and achievements with the following:
\begin{itemize}
    \item a task-aware self-supervised method jointly trained with the learner - MoBYv2;
    \item a quantitative evaluation with MoBYv2AL on multiple image classification benchmarks such as: CIFAR-10/100\cite{cifar}, SVHN\cite{Goodfellow2014svhn} and FashionMNIST \cite{Xiao2017Fashion-MNIST:Algorithms};
    \item state-of-the-art performance over the existing AL baselines.
\end{itemize}

\section{Related Works}
\noindent \textbf{Recent Advances in Active Learning.}
Recent advancement in AL are either uncertainty-oriented \cite{mcdropoutal,Gal2016DropoutGhahramani, Yoo2019LearningLearning, comm, caramalau2021active, Sinha2019VariationalLearning} or data representativeness \cite{Sener2017ActiveApproach, coresvm, cdal};  and  some of them are the mixture of both \cite{BeluchBcai2018TheClassification, csal, kim2021task, ugcn}. 

Under the pool-based setting \cite{settles.tr09}, deep active learning has been initially tackled with uncertainty estimation. 
For classification tasks, this was addressed from the maximum entropy of the posterior or through Bayesian approximation with Monte Carlo (MC) Dropout\cite{mcdropoutal, Gal2016DropoutGhahramani, caramalau2021active}. Concurrently, methods that used latent representations to sample have outperformed the ones that explored uncertainty. From these works, we recognise CoreSet \cite{Sener2017ActiveApproach} as the most revised and competitive baseline.
However, more recently, a new trend shifted the AL acquisition process to parameterised modules. The first work, Learning Loss \cite{Yoo2019LearningLearning} optimises a predictor for the loss of the learner. Still tracking uncertainty, VAAL \cite{Sinha2019VariationalLearning} deploys a dedicated variational auto-encoder (VAE) to adversarial distinguish between labelled and unlabelled images. CoreGCN\cite{ugcn} and CDAL \cite{cdal}, on the other hand, proposed to improve data representativeness with graph convolutional networks and categorical contextual diversity, respectively. We test these methods in the experiments section and we further detail them in the Supplementary.
Given the shared selection criteria with CoreSet, our MoBYv2 AL framework falls in the representativeness-based category.\\
\noindent \textbf{Self-supervised Learning (SSL).}
For the past years, a new pillar, SSL, has arisen in unsupervised environments with linked goals to AL. Learning generalised concepts from large-scale data is critical for further expansion to various vision applications. We can divide the SSL in two approaches: consistency-based \cite{mixmatch, fixmatch, meanteacher, dino} and contrastive energy-based \cite{moco, chen2020mocov2, simclr, byol, CoMatch, xie2021moby}. Consistency regularisation looks to preserve the class of unlabelled data even after a series of augmentations. For example, both MixMatch \cite{mixmatch} and DINO \cite{dino} sharpen the averaged pseudo-labelled predictions. Conversely, contrastive learning generally demands pairs of positive and negative examples while optimising the similarity/contrast between them. Dual networks are usually deployed to evaluate these losses either within the batch (as in SimCLR \cite{simclr}) or within a dictionary of keys (for methods like MoCo\cite{moco}, MoBY\cite{xie2021moby}). Because contrastive learning is foundational to our proposal, we revise these techniques in Sec.~\ref{Sec:method}. \\
\noindent \textbf{AL with self-supervision.}
\label{subsection:alssl}
In the beginning, SSL and AL evolved in parallel. Only recently, these fields have merged to further progress data sampling. Although SSL brings better visual constructs, there is still the question of which labelled information to allocate. By leveraging the unlabelled data behaviour, CSAL \cite{csal} firstly integrated MixMatch in the AL training and selection. We follow a similar strategy, but our end-to-end training learns contrastive representations. Despite this, CSAL is included in the SSL-based experiment as it is directly comparable. Two new works tackle contrastive learning either in acquiring language samples, CAL \cite{calssl}, or by adapting the sequential SSL SimSiam \cite{chen2020simsiam} in \cite{bengar2021reducing}. CAL is task-dependent on natural language processing. In \cite{bengar2021reducing}, the multi-stage AL selection has no effect against random sampling. To this extent, we omit these works in our analysis.
\section{Methodology}
\label{Sec:method}
In this section, we explain our pipeline in detail. First, we introduce deep active learning for image classification in general, followed by our contributions. 

Standard AL requires an online environment where the task learner selects and optimises simultaneously. 
We consider a large unlabelled pool of data $\mathbf{D}_U$ from which we uniformly random sample and label an initial subset $\mathbf{S}^0_L << \mathbf{D}_U$. Let $(\mathbf{x}^L, \mathbf{y}^L) \in \mathbf{S}^0_L$ be the available images and their corresponding classes. Commonly, we deploy a learner by a DL model comprising of a feature encoder $\textbf{f}$ and a class discriminator $\textbf{g}$. The objective loss for the learner is the categorical cross-entropy defined as $\mathcal{L}_{classification} = - \sum_{\mathbf{S}^0_L} \mathbf{y}^L \cdot \log \textbf{g}(\textbf{f}(\mathbf{x}^L))$.

Following the AL objective, we decide upon the \emph{exploration-exploitation trade-off} in conjunction with our classification performance. Thus, we set up the exploitation rate through a budget $b$ across $\mathbf{D}_U /\ \mathbf{S}^0_L$ guided by a \emph{selection criteria}. Consequently, we label the new sampled subset $\mathbf{S}^1_L$ and re-train our learner. The exploration factor is expressed by the \emph{number of stages} $\mathbf{S}^{0 \dots N}_L$  we repeat this loop according to the targeted performance.  While we may limit the exploration cycles, in our proposal, we primarily focus on exploitation. 


\begin{figure*}
    \centering
    \includegraphics[trim=0cm 0cm 0cm 0cm, clip, width=1\textwidth]{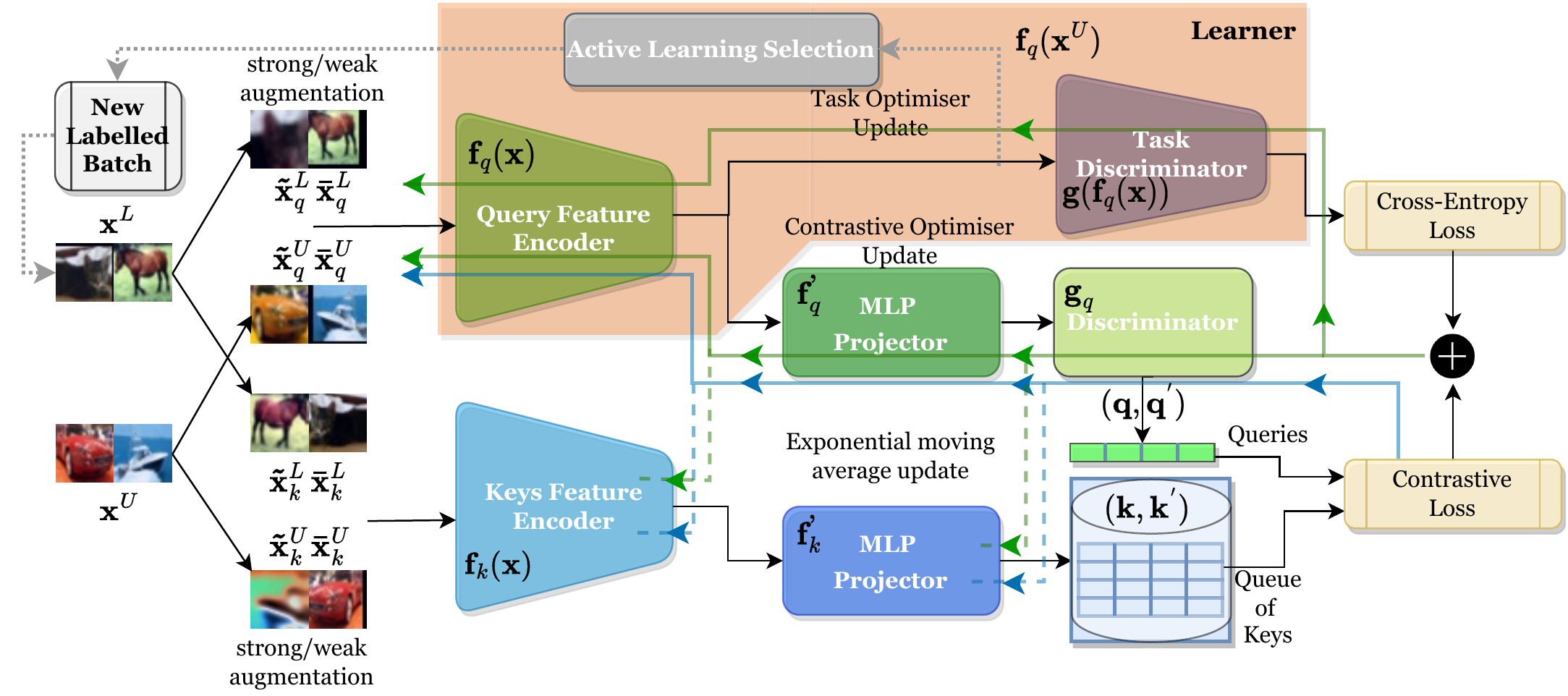}
    \caption{SSL-AL training framework under the proposed MoBYv2AL configuration. The query feature encoder plays two roles: to map the features to the task discriminator for classification; to capture contrastive visual representation with the asymmetry of the query and key modules. For unlabelled data, the \textcolor{blue}{blue lines} show the back-propagation of contrastive loss and its exponential moving average (dashed). The \textcolor{green}{green lines} also include the cross-entropy loss during training when the annotation is available. Once training ends, the unlabelled samples pass through the \textcolor{orange}{learner} for \textcolor{gray}{AL selection}.}
    \label{fig:pipeline}
\end{figure*}
\noindent \textbf{Contrastive Semi-supervised learning framework.} \label{subsec_cssl}
We tackle the contrastive unsupervised learning approach compared to previous semi-supervised AL techniques \cite{csal, revival_iccv21, altod} that rely on consistency measurement. Here, we briefly re-introduce the key aspects of the previous SSL techniques. These are constituent to our \emph{MoBYv2AL} proposal.

The goal of self-supervised learning aligns with the AL problem, where there is plenty of unlabelled data and a costly annotation procedure. However, the former tends to learn generalised visual representation in aid of the objective task. For contrastive learning, the main approach to obtaining these representations is by analysing the similarity (dissimilarity) within the data space. From the most successful works \cite{moco,byol,simclr,dino,xie2021moby}, we can broadly form the contrastive learning process of these main parts: data augmentation with or without dual encoder, feature-vector projections, and similarity approximation by a dedicated loss function.

We design the self-supervision framework according to \emph{MoBY} \cite{xie2021moby}. This method combines two innovative prior works \emph{MoCo}\cite{moco} and \emph{BYOL}\cite{byol} on visual transformers \cite{vi,vt}. 
MoCo\cite{moco} pioneers contrastive learning by addressing the similarity between an image and a specific dictionary of samples. To deploy the loss, positive examples are required through data augmentation of the input query together with the other negative keys from the dictionary. The self-supervision training pipeline consists of two feature encoders and two MLP projectors for mapping the query and the keys. Consequently, the keys are permuted in a large memory bank, while the positive examples are inferred through the online encoder. The gradient over the dictionary of keys needs a slower update. Thus, a gradual momentum update is implemented.

BYOL\cite{byol}, on the other hand, has a different approach for contrastive self-supervision. It simplifies MoCo by relying only on positive examples. In this way, the memory bank can be discarded. The InfoNCE\cite{infonce} loss is also replaced with a \textit{l2} loss given the new setting. The contrastive learning strategy of BYOL is indirectly obtained through batch normalisation. To achieve this, further modifications are proposed. Thus, the architecture of the dual encoders is asymmetric in regard to MoCo, and BYOL adds a prediction module to the projector of the online encoder. Following only positive examples, the inputs to the two networks are strong-augmented versions of the same image. Finally, BYOL preserves the common mode from the data and inherits contrastive learning when passing a slow exponential moving average from the online to the momentum encoder. We intuitively explore the contrastive learning strategies from both MoCo and BYOL and align the self-supervision with MoBY\cite{xie2021moby}. We further present the combined pipeline depicted in \ref{fig:pipeline}.

From a design perspective, we adopt the asymmetric dual encoders from BYOL as shown in Fig.~\ref{fig:pipeline}. The top branch in Figure~\ref{fig:pipeline} culminates with a discriminator $\mathbf{g}_q$ to match the outputs from the bottom. Despite this, both branches consist of the same feature extractor architecture followed by an MLP projector ($\mathbf{f}^{'}_q, \mathbf{f}^{'}_k$ for query and key, respectively). Distinctively from MoBY, we tackle convolutional neural networks (CNNs) as feature encoders. Moreover, we reduce the MLP projectors and the query discriminator to a single layer with batch normalisation and ReLU activation.

The asymmetric pipeline helps to mimic the contrastive learning principle of BYOL. However, to include the concepts from MoCo, we minimise our objective with the InfoNCE loss. In this case, we will also need to keep the memory bank for the queue of keys. We define the contrastive loss as a sum of InfoNCE from two augmented versions of a query $\{q, q{'}\}$ and of a different key $\{k, k{'}\}$:

\begin{equation} \label{eq:1}
\mathcal{L}_{contrastive} = -\log \frac{exp(q \cdot k^{'} \slash \tau)}{\sum_{i=0}^m \exp(q \cdot k_i^{'} \slash \tau)} 
- \log \frac{exp(q^{'} \cdot k \slash \tau)}{\sum_{i=0}^m \exp(q^{'} \cdot k_i \slash \tau))},
\end{equation}
where $m$ is the size of the memory bank and $\tau$ is the adjusting temperature \cite{wu2018temperature}.
During training, the online query encoder branch is updated by gradient while the key encoder takes the slow-moving average with momentum. We ensure with this combined design the preservation of both MoCo and BYOL representation concepts. On the one hand,  the asymmetric structure indirectly finds discrepancies from the average image with moving average and batch normalisation. On the other hand, the contrastive loss with the queue of different keys maintains the direct distinctiveness between the images.

The standard SSL techniques MoCo, BYOL and MoBY demand the supervision stage where the pre-trained models are fine-tuned for the task objective. Such multi-stage pipelines seem ineffective
in AL~\cite{bengar2021reducing}. In this paper, we extended the SSL pipeline of \emph{MoBY} 
to minimise both the self-supervised objective and downstream task objective jointly. \\
\noindent \textbf{Joint Objective.}
A final step to clarify before presenting the joint training procedure is data augmentation. MoBY derives the augmentation strategy from BYOL, where the inputs suffer strong transformations. In our proposal, we choose an alternation between strong and weak augmentation, similarly to MoCov2\cite{chen2020mocov2}. This change boosted the performance of its predecessor \cite{moco}. We also observed in our experiments that using only strong augmentations can affect the optimisation of the task-aware branch. The weak augmentations comprise horizontal flips and random crops. In addition, the strong transformation includes colour jitter (on brightness, contrast, saturation, hue), Gaussian blur, grayscale conversion and pixel inversion (solarise). From equation \ref{eq:1}, $\{q, k\}$ can be referred as the weak transformations of query and key, and $\{q^{'}, k^{'}\}$ their corresponding stronger versions.

With all these elements in place, we can change the learner from the existing AL framework with the modified MoBY and train jointly the pipeline. Starting from the first cycle, we consider the available labelled samples $(\mathbf{x}^L, \mathbf{y}^L) \in \mathbf{S}^0_L$ and the remaining unlabelled $\mathbf{x}^U \in \mathbf{D}_U$ as queries and keys. A strong augmentation is marked as $\{\mathbf{\tilde x}^L_q, \mathbf{\tilde x}^L_k\}$, while a weak is represented with $\{\mathbf{\bar x}^L_q, \mathbf{\bar x}^L_k\}$. When training, we alternate between batches of labelled and unlabelled data with every inference. Therefore, we back-propagate only the contrastive loss for the unlabelled to \ref{eq:1}. In this context, given the pipeline from Figure \ref{fig:pipeline} for this contrastive loss $\mathcal{L}^U_{contrastive}(q,q^{'};k,k^{'})$, $\{q, k\}$ and $\{q^{'}, k^{'}\}$ can be obtained so:

\begin{gather} \label{eq:2}
    \{q, q^{'}\} = \mathbf{g}_q \:( \mathbf{f}^{'}_q \:( \mathbf{f}_q( \:\{\mathbf{\bar x}^U_q , \mathbf{\tilde x}^U_q \}))), \\
    \{k, k^{'}\} = \mathbf{f}^{'}_k \:( \mathbf{f}_k \:(\{\mathbf{\bar x}^U_k, \mathbf{\tilde x}^U_k \})).
\end{gather}

Similarly, we can compute $\mathcal{L}^L_{contrastive}$, the contrastive loss for the labelled images. In addition, we also minimise the categorical cross-entropy, $\mathcal{L}_{classification}$, with the output from the task discriminator. Once computed, we back-propagate both the contrastive and the classification loss. Therefore, the combined loss, adjusted by a scaling factor $\lambda_c$, can be expressed as:

\begin{equation} \label{eq:3}
\mathcal{L}^L_{combined} =  \mathcal{L}_{classification} + \lambda_c \mathcal{L}^L_{contrastive}
\end{equation}
While the contrastive loss is computed continuously regarding the classification loss, we decide to reduce its influence over the gradients with $\lambda_c = 0.5$.
Finally, it is worth mentioning that the exponential moving average and the queue of keys are updated on the bottom branch for both labelled and unlabelled samples. \\
\noindent \textbf{Unlabelled samples selection.} \label{sub_section:ssel}
We emphasise that our proposal minimises the self-supervised loss inspired by MoBY. With this, the end-task objective jointly enriches the visual representations of the data compared to the standard AL strategy. AL selection methods that rely on the learner's data distribution will perform better. 
CoreSet \cite{Sener2017ActiveApproach} has been proven to be effective in such scenario. To this extent, we primarily choose this selection function with MoBYv2AL. Briefly, CoreSet aims to find a subset of data points where a constant radius bounds the loss difference with the entire data space. This technique is approximated with k-Centre Greedy algorithm \cite{wolf} in the euclidean space of our feature encoder outputs $\mathbf{f}_q(\mathbf{x})$. A thorough visual selection of different AL selection approaches together with CoreSet in presented in the Supplementary.


\section{Experiments}

\textbf{Datasets.}For the quantitative evaluation, we put forward four well-known image classification datasets: CIFAR-10, CIFAR-100 \cite{cifar}, SVHN\cite{Goodfellow2014svhn} and FashionMNIST\cite{Xiao2017Fashion-MNIST:Algorithms}. 

\noindent \textbf{Models.} We mentioned in \ref{subsec_cssl} that we use different CNNs for feature encoders. To show that MoBYv2AL is robust to architectural changes, we opt for VGG-16 \cite{Simonyan2015VeryRecognition} in the CIFAR-10/100 quantitative experiments and for ResNet-18 \cite{he2016deep} in SVHN and FashionMNIST.

\noindent \textbf{Training settings.} 
We train at every selection stage for 200 epochs, and we keep the batch size at 128. The dictionary size for the keys $m$ is set up as in MoBY at 4096. We noticed in our experiments that the contrastive and cross-entropy loss converge together after 200 epochs. The learning rate starts at 0.01, and it follows a schedule for the queue encoder and task discriminator that decreases ten times at 120 and 160 epochs. However, we keep the momentum scheduler update in the key bottom branch (gradual momentum increment from 0.99). In the contrastive loss, for both queues, we fix the temperature parameter to 0.2.

\noindent \textbf{AL settings.} We followed the AL settings of VAAL\cite{Sinha2019VariationalLearning}, CDAL\cite{cdal} and CoreGCN\cite{ugcn}. For more details, please see Supplementary. 

\noindent \textbf{Baselines.} 
We compared our method MoBYv2AL with a wide range of methods in active learning such as: 
MC Dropout \cite{mcdropoutal}, DBAL \cite{Gal2016DropoutGhahramani}, Learning Loss\cite{Yoo2019LearningLearning}, VAAL\cite{Sinha2019VariationalLearning}, , Learning Loss~\cite{Yoo2019LearningLearning}, CoreGCN\cite{ugcn} and CDAL\cite{cdal}. 


\subsection{Quantitative experiments}
\label{subsec:quant_exp}
\begin{figure*}
    \centering
    \includegraphics[trim=0cm 0cm 0cm 0cm, clip, width=.44\textwidth]{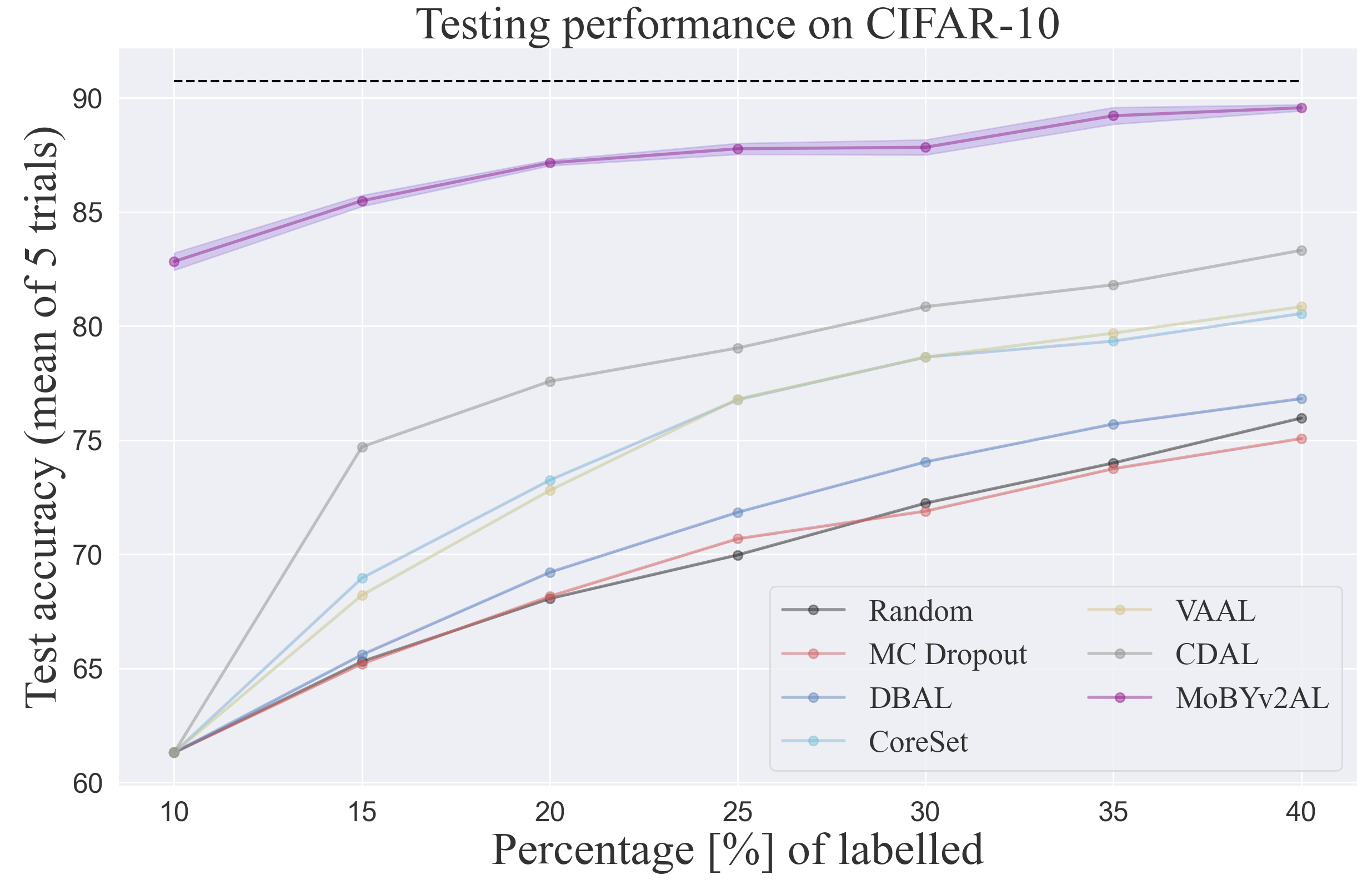}
    \includegraphics[trim=0cm 0cm 0cm 0cm, clip, width=.44\textwidth]{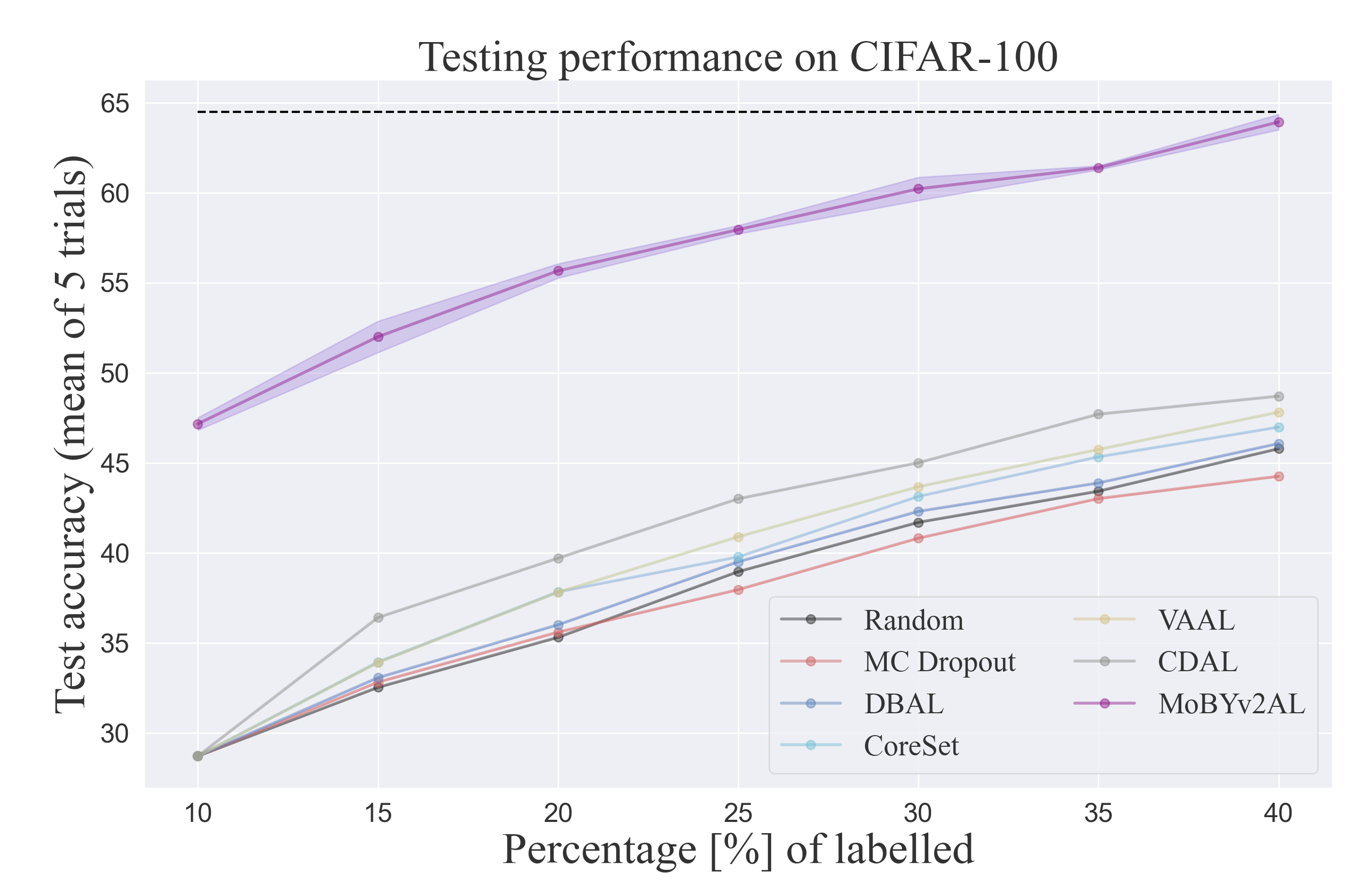}
    \caption{Evaluations on CIFAR-10 (\textbf{left}), CIFAR-100 (\textbf{right}) [Zoom in for better view]}
    \label{fig:cifar10/0}
\end{figure*}

\textbf{CIFAR10/100.} To maintain a fair comparison, in Figure \ref{fig:cifar10/0}, we report the performance charts obtained by CDAL\cite{cdal} and VAAL\cite{Sinha2019VariationalLearning}. All methods use VGG-16 for the feature encoder. MoBYv2AL has a considerable advantage with the proposed SSL framework in the CIFAR-10/100 experiments from the first selection stage. In both scenarios, we gain 20\% testing accuracy over standard learning (62\% and 28\% on CIFAR-10/100). This justifies the importance of the joint training framework from MoBYv2AL.

Our pipeline's more refined visual representations direct helpful information to the CoreSet selection method. Thus, we notice a gradual increase in Figure \ref{fig:cifar10/0}, where after 7 cycles, with 40\% labelled data, MoBYv2AL achieves 89.6\% mean accuracy on CIFAR-10 and 63.1\% on CIFAR-100. Another observation in the CIFAR-10 experiment is that the AL performance saturates faster than in CIFAR-100. This effect occurs due to a large initial labelled pool in relation to the complexity of the task. MoBYv2 exploits more contrastive information, and it limits the exploratory potential in the next stages.

\begin{figure*}
    \centering
    \includegraphics[trim=0cm 0cm 0cm 0cm, clip, width=.44\textwidth]{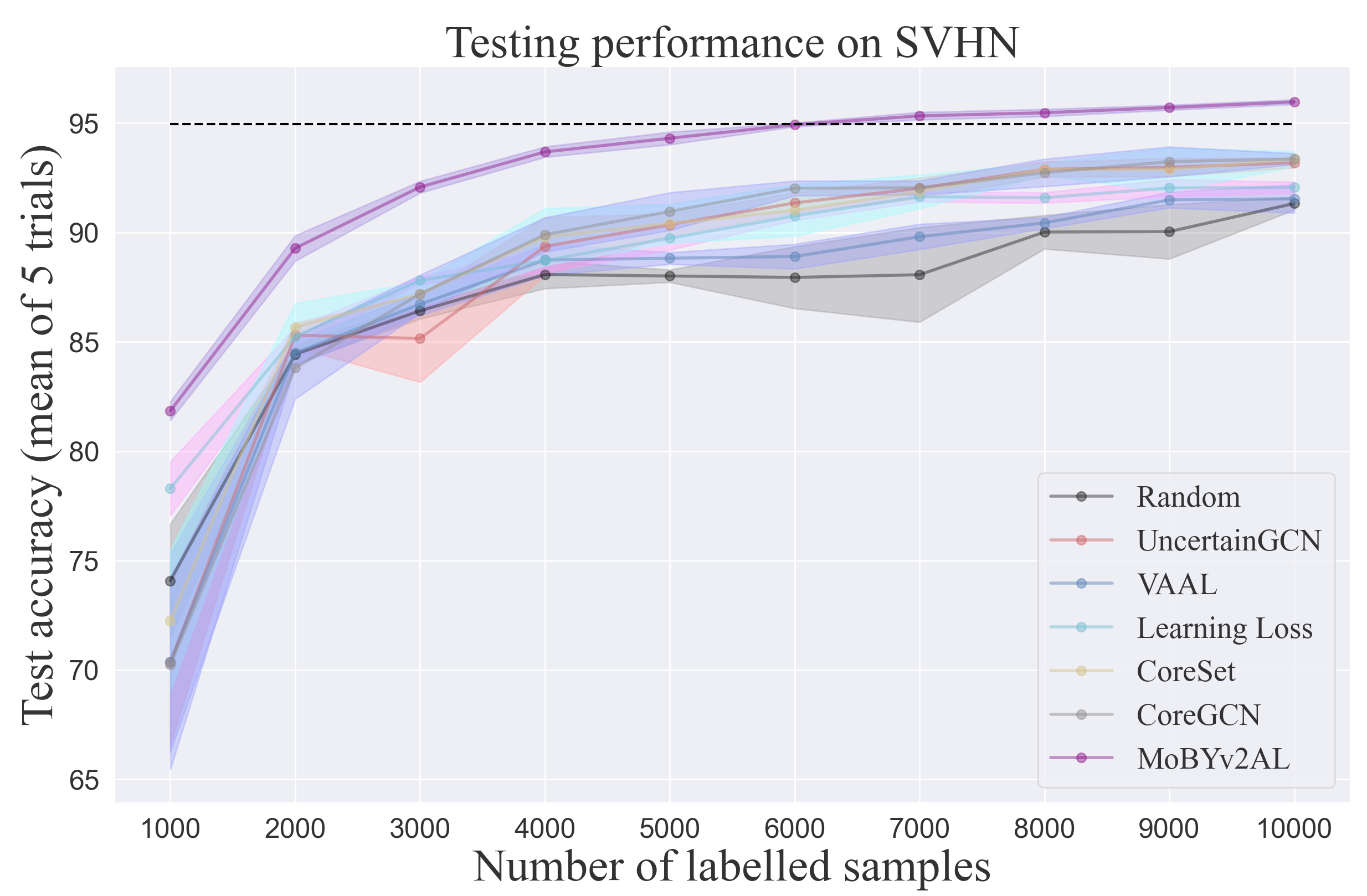}
    \includegraphics[trim=0cm 0cm 0cm 0cm, clip, width=.44\textwidth]{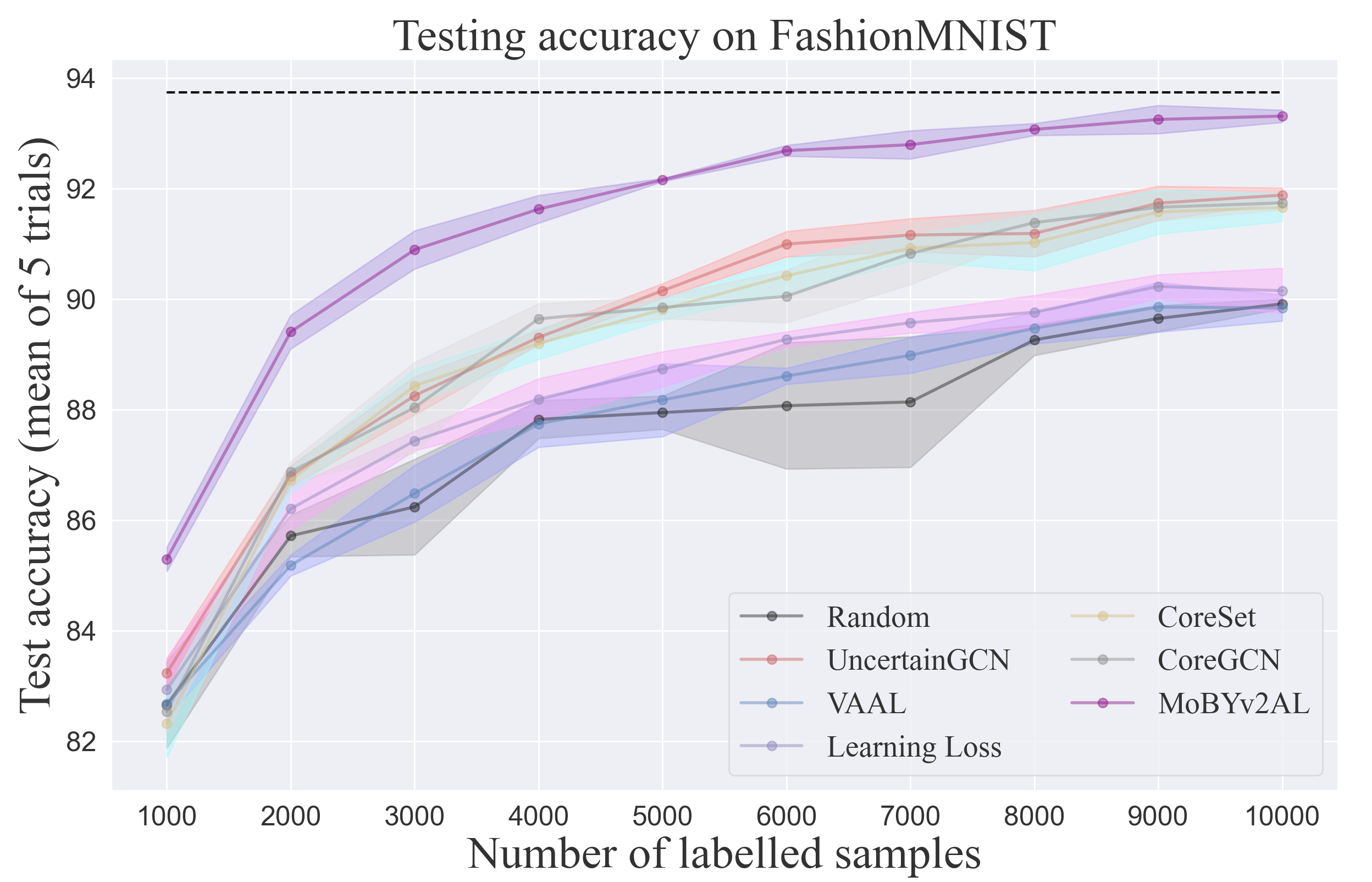}
    \caption{Evaluations on SVHN (\textbf{left}), FashionMNIST (\textbf{right}) [Zoom in for better view]}
    \label{fig:fmnist}
\end{figure*}
\noindent \textbf{SVHN/FashionMNIST.} We can deduct, from Figure \ref{fig:fmnist} as well, that MoBYv2AL balances the exploration-exploitation trade-off when the initial labelled set is relatively low to the number of classes.
The dark dashed line displays the supervised baseline training on the entire labelled set. While on CIFAR-10/100 and FashionMNIST, MoBYv2AL reaches comparable performance, by the end of the cycles, on SVHN, it surpasses after the sixth one (95\%). Here, we emphasise the relevance of the strong/weak augmentations in enriching the discrete data distribution. Furthermore, grayscale data (as in FashionMNIST) can also benefit from the proposed AL framework. In Figure \ref{fig:fmnist}, we keep the same results of the previous baselines from CoreGCN\cite{ugcn}. Even under these settings, we outperform the state-of-the-arts with a noticeable consistent margin: for SVHN and FashionMNIST a gap of at least 2\% - 3\%. 

\begin{table}[hbt!]

\centering
\begin{tabular}{l|l|l|l|l|l}
SSL-AL method vs percentages of labelled & 10\%           & 15\%           & 20\%           & 25\%    & 30\%         \\ \hline
CSAL       & 58.1          & 63.76           & 67.13          & 69.28  &70.08          \\
MoBYv2AL   & \textbf{67.66}   & \textbf{68.24} & \textbf{68.49} & 68.57 & \textbf{70.11}
\end{tabular}
\caption{Comparison with the SSL-AL method CSAL on CIFAR-100 with a WideResNet-28 learner}
\vspace{-4mm}
\label{tab:csal}
\end{table}
\noindent \textbf{Comparison with other SSL-AL.} MoBYv2 leverages unlabelled data for contrastive learning in the AL framework. Previously, we chose this amount of data equal to the available labelled samples. Therefore, at every AL cycle, this size increases with the newly selected data. Another recent SSL-AL baseline CSAL\cite{csal}, however, deployed the consistency measurements from MixMatch\cite{mixmatch} on the entire unlabelled data. We could identify that MoBYv2AL over-exploits as CSAL the captured representation under these conditions. We further compare the 2 methods on CIFAR-100 in Table \ref{tab:csal} and adjust the feature encoder to WideResNet-28\cite{wideresnet}. In this experiment, MoBYv2AL maintains the initial performance gain.

\begin{figure*}[hbt!]
    \centering
    \includegraphics[trim=0cm 0cm 0cm 0cm, clip, width=.44\textwidth]{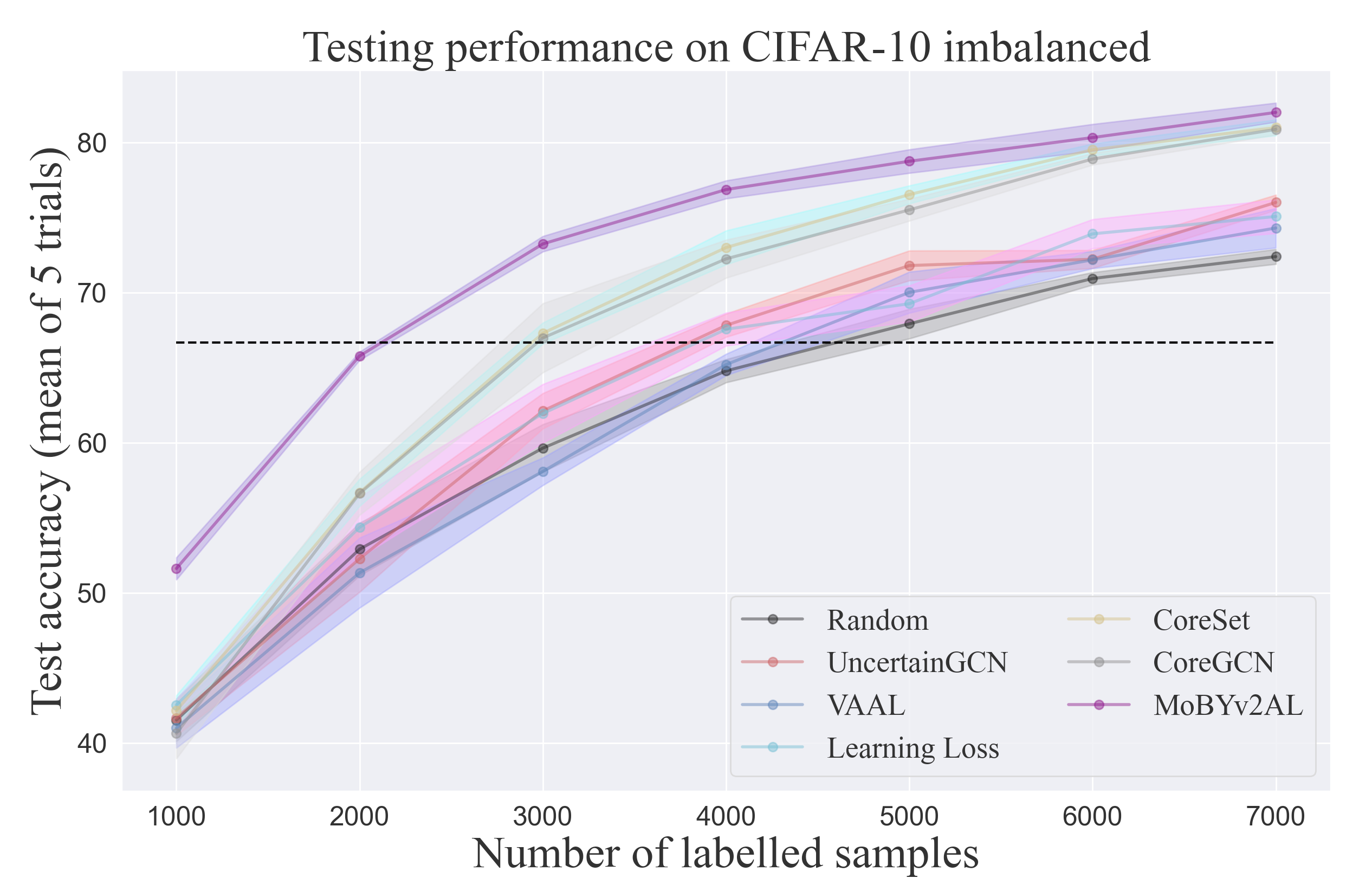}
    \includegraphics[trim=0cm 0cm 0cm 0cm, clip, width=.46\textwidth]{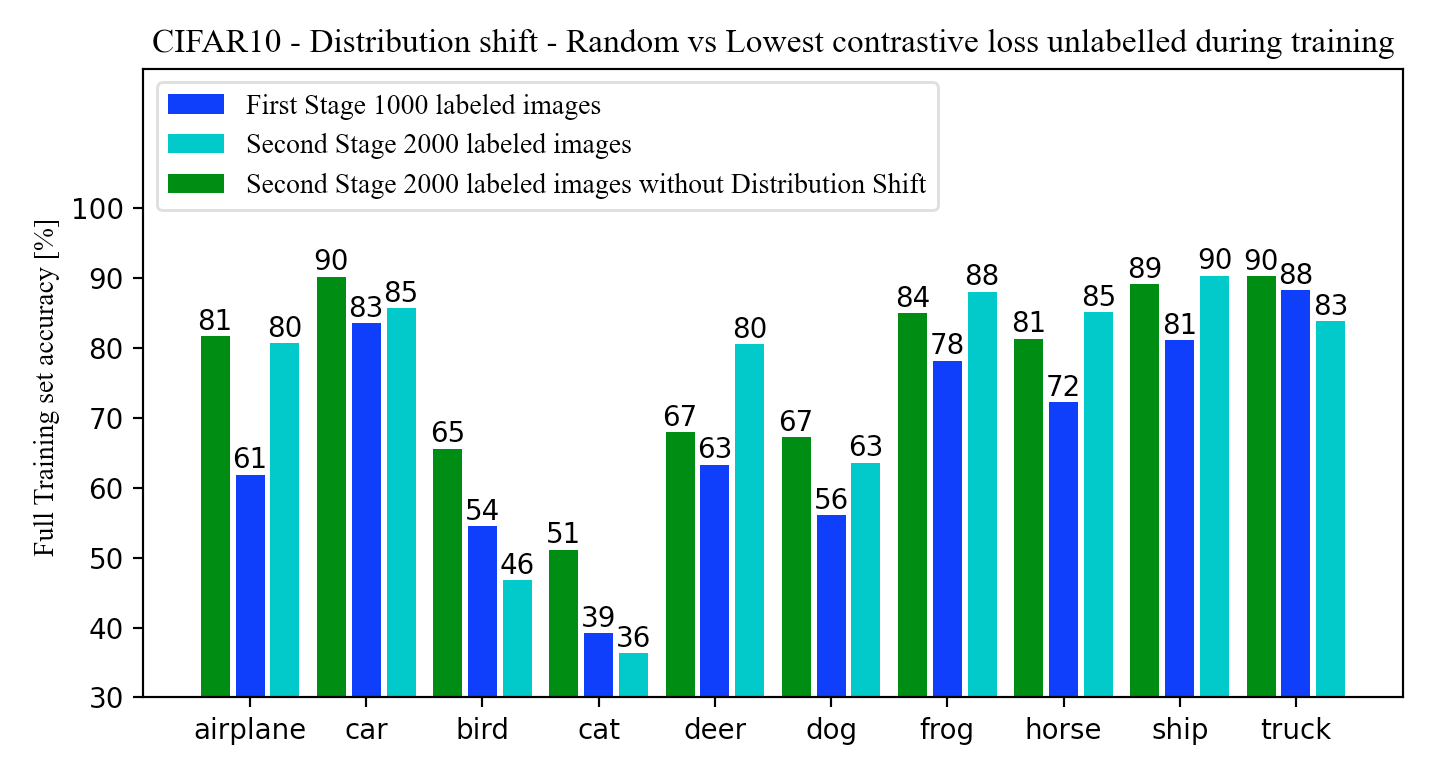}
    \caption{CIFAR-10 imbalanced dataset experiment(\textbf{left}); Mitigating the distribution shift with MoBYv2AL(\textbf{right}) [Zoom in for better view]}
    \label{fig:imbdist}
\end{figure*}

\noindent\textbf{Imbalanced dataset experiment.}
Apart from SVHN, all the previous experiments have a uniform distribution over the classes. This rarely occurs during real-world acquisition scenarios. Therefore, as in CoreGCN, we simulate an imbalanced CIFAR-10 unlabelled set. Each of the ten classes has originally 5000 training examples. We decide to reduce 5 of the classes to 500 images (resulting in a pool of 27500). The learner contains a ResNet-18 encoder, and it is trained with an initial set of 1000 labelled examples. We apply MoBYv2AL together with the other baselines from CoreGCN\cite{ugcn} for 7 cycles. Figure \ref{fig:imbdist}(left) presents the ability of MoBYv2AL to outperform the previous methods even in possible real-world environments. Investigation of long-tail distributions is still part of our future work. 

\subsection{Distribution shift discussion}
In deep AL, the cyclical process of re-training the learner with the new labelled data may result in optimising to different local minima. Therefore, the exploration and exploitation of the AL method will be affected by this distribution shift at every stage. During experiments, this is commonly shown through jaggy curves (especially for uncertainty-based methods like MC Dropout\cite{mcdropoutal}, DBAL\cite{Gal2016DropoutGhahramani} or UncertainGCN\cite{ugcn}). To address this known issue \cite{Kirsch2021TestDA}, we analyse MoBYv2AL performance on the entire CIFAR-10 training set when providing 1000 and 2000 samples.

The dark blue bars of each class in Figure \ref{fig:imbdist} (right) level the corresponding classification accuracy with the first 1000 random samples. Tracking the performance on the entire set challenges the learner to prefer certain classes. We continue to select with MoBYv2AL another set of images. Consequently, the resulted accuracy is displayed by the cyan bar. We can clearly observe that the minima shifted in a different direction where only some classes improved at the expense of the others. To mitigate this shift, we investigated what impact the unlabelled samples have in our end-to-end training. These samples play a key role in building up the dictionary of keys. Our insight is that the CoreSet selection on MoBYv2 data representation targets primarily high contrastive samples. We can control this effect by customising the unlabelled set deployed in training our learners. To this extent, we propose to use the unlabelled data with the lowest contrastive loss. In Figure \ref{fig:imbdist} (right), we displayed on green bars the performance with this mechanism. From an initial 1000 set accuracy (dark blue) we get an effective linear increase for all the 10 classes. This effect is consistent throughout all the previous quantitative experiments as well.
\subsection{SSL modules variation and ablation study}
We continue to motivate the proposed design of MoBYv2AL with a set of ablation experiments and by varying its SSL module. On the left side of Table \ref{tab:abl}, we swap in the end-to-end training pipeline the original version of MoBY \cite{xie2021moby} and the preceding SSL state-of-the-arts, MoCov2 \cite{chen2020mocov2} and BYOL \cite{byol}. Apart from MoBY, the learner did not converge on any selection cycle with the other SSL modules. Thus, the setup of large batches and specific training conditions (low learning rates, cosine scheduler) and learners can hardly adapt to this semi-supervision configuration.
For MoBYv2AL, the weak-augmented inferences to the learner stabilise its performance in regard to the original version. Furthermore, our method distances by 4\% class accuracy with each AL cycle. 
\begin{table}[hbt!]
\scalebox{0.65}{
\begin{tabular}{l|c|c|c}
\multicolumn{1}{l|}{SSL model / No. of labelled}  &1000 & 2000 & 3000 \\ \hline
MoCov2                      & 11.62±.9                 & 11.92±.6                 & 12.89±.6                 \\ \hline
BYOL                        & 12.32±.7                 & 11.72±.4                 & 11.47±.2                 \\ \hline
MoBY                        & 62.62±.4                 & 72±.5                    & 76.43±.1                 \\ \hline
MoBYv2AL (Ours)               & \textbf{63.06±.5}        & \textbf{76.04±.6}        & \textbf{80.63±.3}       
\end{tabular}}
\quad
\scalebox{0.65}{
\begin{tabular}{c|c|c|c}
\multicolumn{1}{l|}{MoBYv2AL  / No. of labelled} & 1000              & 2000              & 3000              \\ \hline
w/o Discriminator                              & 60.44±.4          & 72.53±.8          & 77.89±.3          \\ \hline
w/o MLP Projector                              & 58.57±.6          & 71.96±.5          & 77.02±.6          \\ \hline
w/o Strong Augmentation                        & 47.7±.4           & 58±.5             & 64.85±.5          \\ \hline
MoBYv2AL (Ours)                                         & \textbf{63.06±.5} & \textbf{76.04±.6} & \textbf{80.63±.3}
\end{tabular}}
\caption{Variation of SSL pipeline (\textbf{left}) and ablation study of MoBYv2AL (\textbf{right}). Average testing performance (5 trials) on CIFAR-10 for 3 AL cycles with ResNet-18 encoder}
\label{tab:abl}
\end{table}
One can argue that our SSL framework comprises several building blocks, and its implementation can deter developers. While we value the significant dominance of MoBYv2 in AL selection, we still motivate the relevance of each part in Table \ref{tab:abl} (right). In the ablation evaluation, we successfully remove the queue Discriminator and the MLP projectors. As a result, we detect a continuous accuracy drop. Projecting larger features and simulating the asymmetry brings the advantage of contrastive learning in MoBYv2. Moreover, strong augmentations also play a crucial role in the SSL  pipeline.

\begin{table}[hbt!]
\scalebox{0.6}{
\begin{tabular}{c|c|c|c}
\multicolumn{1}{l|}{MoBYv2AL} & 1000     & 2000     & 3000     \\ \hline
Multi-stage semi supervised & 34.8±.1  & 34.96±.2 & 35.09±.1 \\ \hline
Jointly with end-task       & 63.06±.5 & 76.04±.6 & 80.63±.3
\end{tabular}}
\scalebox{0.55}{
\begin{tabular}{l|c|c|l|l|l}
SSL method                                  & Supervised & MoCov2 & BYOL  & DINO & MoBYv2AL         \\ \hline
\multicolumn{1}{c|}{CIFAR-10 Test accuracy} & 90.08      & 76.7   & 77.89 & 81.2 & \textbf{88.62}
\end{tabular}}
\caption{Multi-stage SSL-AL vs Jointly end-task AL (\textbf{left}). Semi-supervised learning comparison (\textbf{right}). Testing performance on CIFAR-10 with ResNet-18 encoder}
\label{tab:abl_ssl}
\end{table}

\subsection{SSL results and multi-stage AL}
MoBYv2 SSL for AL strategy is designed in a joint manner with the end task. Despite this, the recent work \cite{bengar2021reducing} that proposes contrastive learning with SimSiam\cite{chen2020simsiam} adopts multi-stage learning for the learner. The pipeline proposed fails to sample better than random in the AL paradigm. In Table \ref{tab:abl_ssl}(left), we experiment with MoBYv2 the multi-stage training (with unsupervised contrastive learning and second task fine-tuning) for CIFAR-10. We observe that the performance suffers in context to the end-task, where limited labelled examples are used. Similarly to \cite{bengar2021reducing}, we also notice a minor improvement when adding more selected data with CoreSet. To this extent, we decided to use the entire training set during fine-tuning. We re-iterated the same experiment for SSL cross-validation with MoCov2\cite{chen2020mocov2}, DINO\cite{dino} and BYOL\cite{byol}.
\section{Limitations and Conclusions}
Although we can adapt MoBYv2AL to other applications, we expect further research on the effects of the augmentations and the momentum encoder. Another limiting factor should be analysed at the first AL selection stage, where developers may tune the exploration-exploitation ratio to avoid saturation.

We have presented an SSL-based  AL framework for image classification. The main contributions lie in the task-aware contrastive learning pipeline. MoBYv2AL retains the higher visual concepts and aligns them with the downstream task. The joint training is efficient and modular, allowing diverse backbones and sampling functions. We conduct quantitative experiments and demonstrate the state-of-the-art on four datasets. Our method shows robustness even in simulated class-imbalanced data pools.

\section{Acknowledgements}
This work is in part sponsored by KAIA grant (22CTAP-C163793-02, MOLIT), NST grant (CRC 21011, MSIT), KOCCA grant (R2022020028, MCST) and the Samsung Display corporation.
BB and DS are funded in whole, or in part, by the Wellcome/EPSRC Centre for Interventional and Surgical Sciences (WEISS) [203145/Z/16/Z]; the Engineering and Physical Sciences Research Council (EPSRC) [EP/P027938/1, EP/R004080/1, EP/P012841/1]; and the Royal Academy of Engineering Chair in Emerging Technologies Scheme; and EndoMapper project by Horizon 2020 FET (GA 863146).
\bibliography{egbib}
\clearpage





\appendix
\counterwithin{figure}{section}
\section{Detailed settings for the AL experiments on MoBYv2AL}
\label{subsec:sup_set}
\textbf{Datasets.}For the quantitative evaluation, we put forward four well-known image classification datasets: CIFAR-10, CIFAR-100 \cite{cifar}, SVHN\cite{Goodfellow2014svhn} and FashionMNIST\cite{Xiao2017Fashion-MNIST:Algorithms}. CIFAR-10 and CIFAR-100 contain the same 50000 training examples but with different labelling systems (10 and 100 classes). SVHN and FashionMNIST are separated into ten classes each as CIFAR-10. However, both datasets are larger, with 73257 coloured street numbers and 60000 grayscale images for FashionMNIST. Although CIFAR-10/100 and FashionMNIST have class-balanced data, this is not the case for SVHN. From another perspective, deploying grayscale images from FashionMNIST challenges our contrastive learning approach, previously customised to RGB data. 

\noindent \textbf{Models.} We mentioned in the Methodology that we use different CNNs for feature encoders. To show that MoBYv2 is robust to architectural changes, we opt for VGG-16 \cite{Simonyan2015VeryRecognition} in the CIFAR-10/100 quantitative experiments and for ResNet-18 \cite{he2016deep} in SVHN and FashionMNIST. Moreover, for the SSL comparison with CSAL we align the encoder with WideResNet-28\cite{wideresnet}.

\noindent \textbf{AL settings.} Under the exploration-exploitation trade-off, we characterise the budget to select as an exploiting factor while the exploration is captured in the number selection cycles. The initial random-sampled labelled dataset varies between the CIFAR-10/100 experiments and SVHN/FashionMNIST. For CIFAR-10/100, we consider  10\% (5000) of the entire training set as labelled and the rest as unlabelled data. The budget is limited to 5\% (2500) samples for selection, and we repeat this cycle seven times. In the second set of experiments, we test our method in a more restrictive environment with a starting set of 1000 labelled and a similar fixed budget. Despite this, we expanded the exploration to 10 cycles reaching 10000 labelled data. As a performance measurement, we evaluate the average of 5 trials testing accuracy in the AL framework.

\section{Selection function analysis} \label{subsection:abl}
\begin{figure*}[hbt!]
    \centering
    \includegraphics[trim=0cm 0cm 0cm 0cm, clip, width=.49\textwidth]{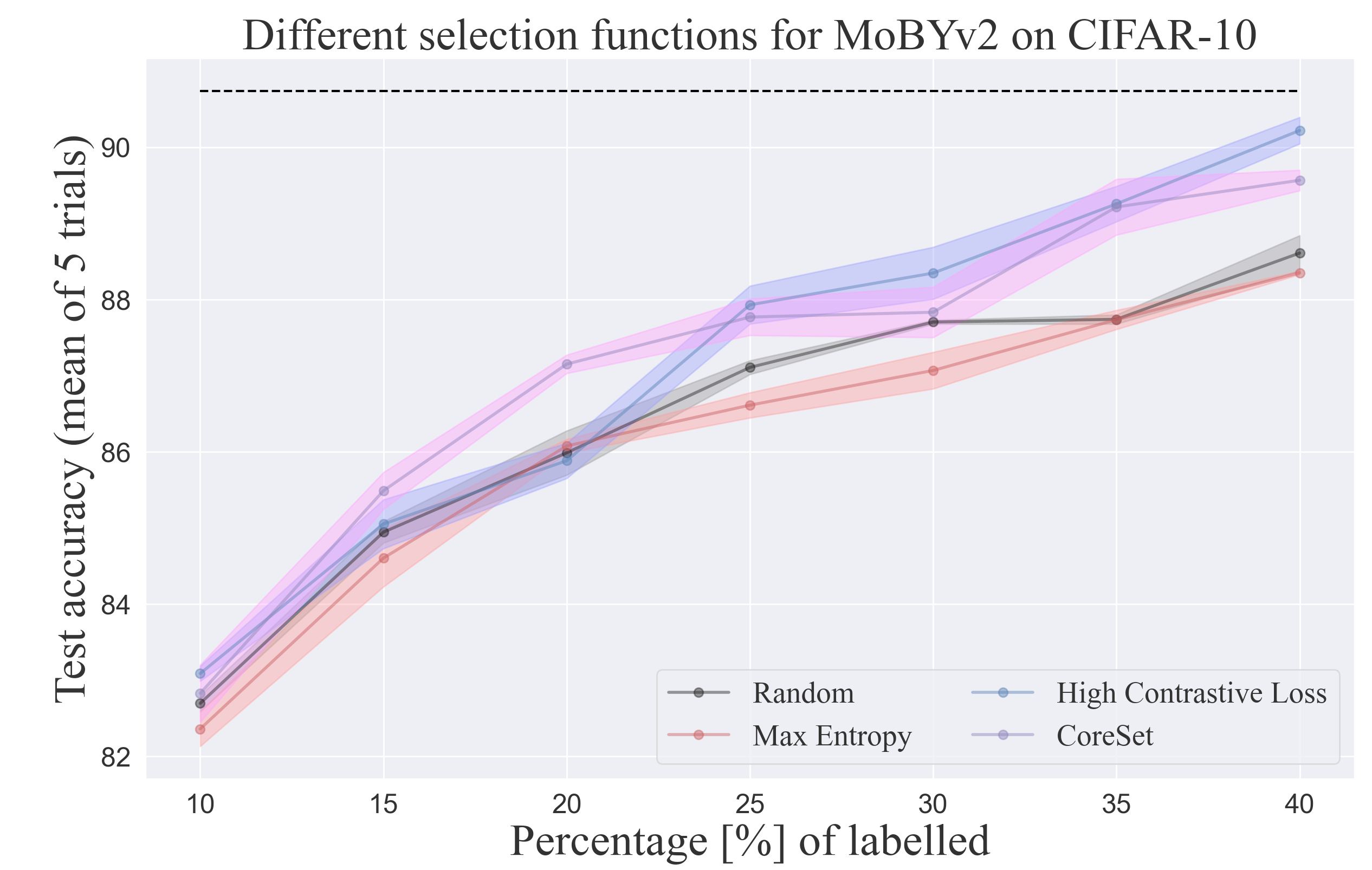}
    \includegraphics[trim=0cm 0cm 0cm 0cm, clip, width=.49\textwidth]{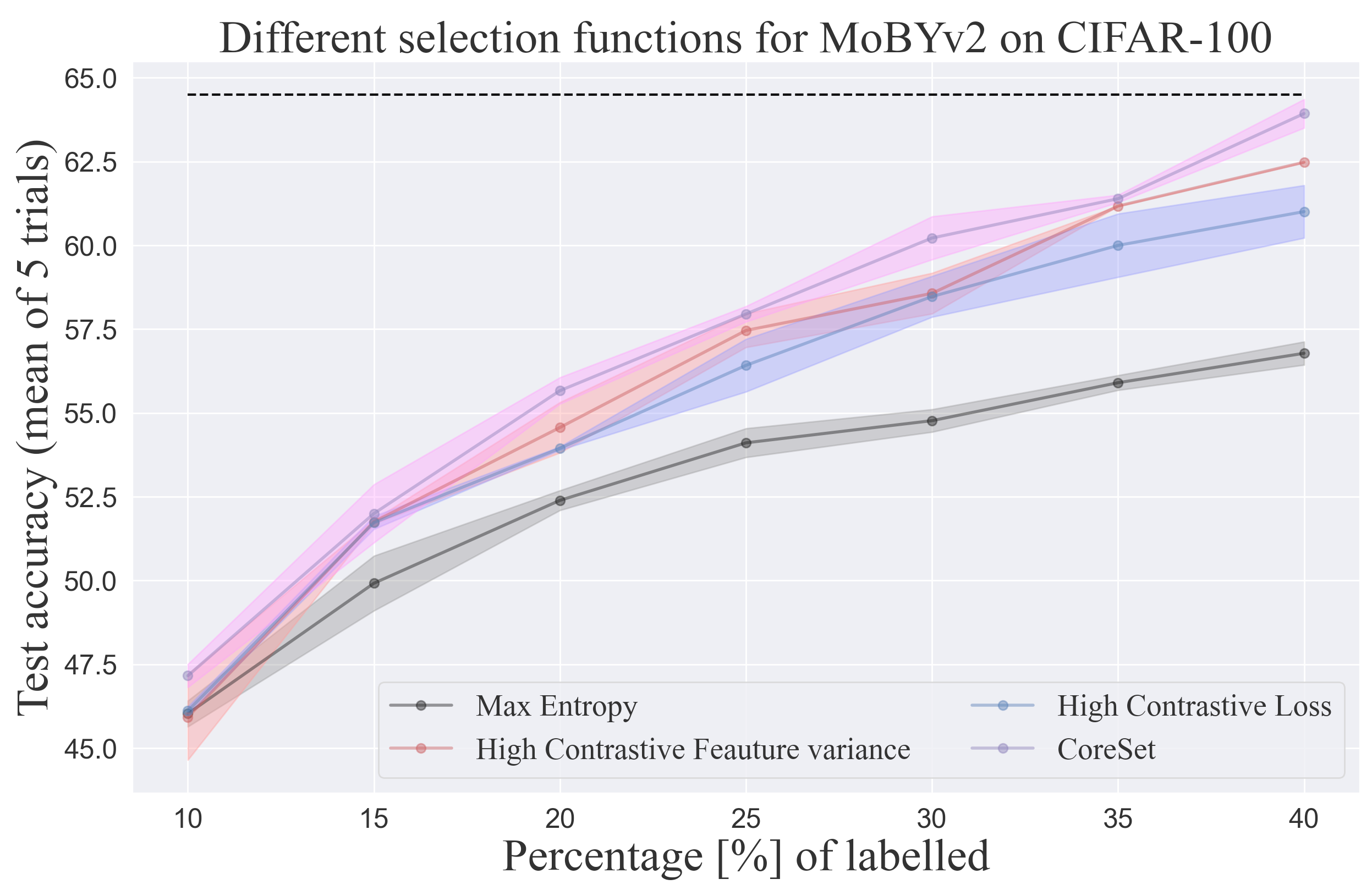}
    \caption{Quantitative evaluation with different selection functions for CIFAR-10 (\textbf{left}), CIFAR-100 (\textbf{right}) [Zoom in for better view]}
    \label{fig:acq}
\end{figure*}
Our proposed pipeline, MoBYv2AL, can easily adapt to multiple selection methods. Here, we quantitatively motivate the choice of CoreSet from section \ref{sub_section:ssel}. Therefore, we re-evaluate MoBYv2AL on CIFAR-10/100 benchmarks in Figure \ref{fig:acq}. We vary the selection of the new budget between random, maximum class entropy and CoreSet. Intuitively, we also analyse the effect of selecting unlabelled examples with high contrastive loss.

In both benchmarks, sampling with random or max entropy benefits the less MoBYv2AL pipeline. On the other hand, a representativeness-oriented method like CoreSet suits our hypothesis better. When sampling with high contrastive loss, we detected repetitive examples from some specific classes. This can be explained by higher contextual variance in that category. Specifically, on CIFAR-10, animal classes (cat, deer, dog), with stronger patterns, were more preferred than the vehicle ones (car, truck, ship).

\begin{figure*}[hbt!]
    \centering
    \includegraphics[trim=0cm 0cm 0cm 0cm, clip, width=.97\textwidth]{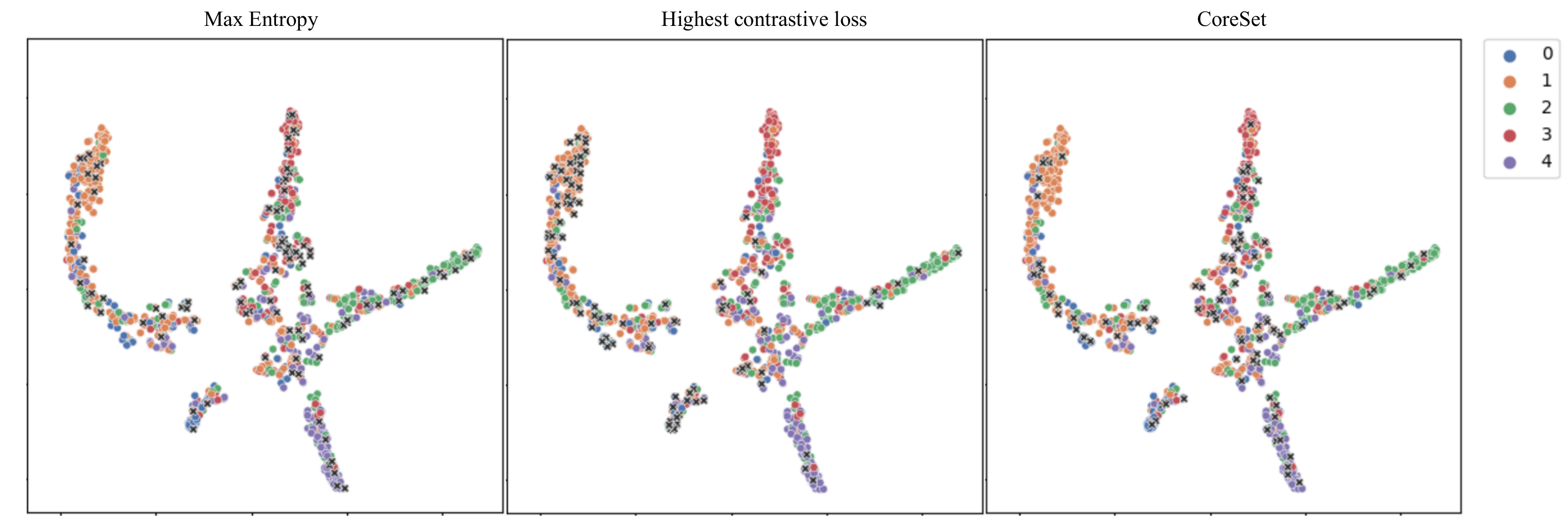}
    \caption{Qualitative AL selection analysis on MoBYv2. t-SNE representations at the first selection stage for 5 classes of SVHN.  [Zoom in for better view]}
    \label{fig:tsne_sel}
\end{figure*}

For a better visual analysis, we have simulated a toy-set experiment with the first five classes from SVHN. Here, we take t-SNE\cite{Maaten08visualizingdata} representations of the MoBYv2AL query encoder outputs of unlabelled data. In Figure \ref{fig:tsne_sel}, the samples marked with crosses construct the new labelled set.
The selection behaviour of the Max Entropy and CoreSet can be interpreted as expected: on the left side, the uncertainty-based technique tracks the most class-variant images; CoreSet, on the right side, samples both in and out-of-distribution according to the Euclidean space.

\end{document}